\documentclass[sigconf]{acmart} 

\AtBeginDocument{%
  \providecommand\BibTeX{{%
    \normalfont B\kern-0.5em{\scshape i\kern-0.25em b}\kern-0.8em\TeX}}}

\setcopyright{acmcopyright}
\copyrightyear{2020}
\acmYear{2020}
\acmDOI{XX.XXXX/XXXXXXX.XXXXXXXX}

\acmConference[]{ACM Conference}{}{}
\acmBooktitle{ACM Conference}
\acmPrice{}
\acmISBN{}

\usepackage{multirow}

\usepackage{amsmath}
\usepackage{bm}
\usepackage{upgreek}
\usepackage{xcolor}

\begin{document}

\title{mEBAL: A Multimodal Database for Eye Blink Detection and Attention Level Estimation}


\author{Roberto Daza, Aythami Morales, Julian Fierrez, Ruben Tolosana}
\affiliation{%
  \institution{Biometrics and Data Pattern Analytics, BiDA-Lab, Universidad Autonoma de Madrid, Spain}}
  %\streetaddress{Anonymous}
  %\city{Spain}}
\email{{roberto.daza, aythami.morales, julian.fierrez, ruben.tolosana}@uam.es}

\renewcommand{\shortauthors}{R. Daza, A. Morales, J. Fierrez, R. Tolosana}

\begin{abstract}
This work presents mEBAL\footnote{\url{https://github.com/BiDAlab/mEBAL}}, a multimodal database for eye blink detection and attention level estimation. The eye blink frequency is related to the cognitive activity and automatic detectors of eye blinks have been proposed for many tasks including attention level estimation, analysis of neuro-degenerative diseases, deception recognition, drive fatigue detection, or face anti-spoofing. However, most existing databases and algorithms in this area are limited to experiments involving only a few hundred samples and individual sensors like face cameras. The proposed mEBAL improves previous databases in terms of acquisition sensors and samples. In particular, three different sensors are simultaneously considered: Near Infrared (NIR) and RGB cameras to capture the face gestures and an Electroencephalography (EEG) band to capture the cognitive activity of the user and blinking events. Regarding the size of mEBAL, it comprises 6,000 samples and the corresponding attention level from 38 different students while conducting a number of e-learning tasks of varying difficulty. In addition to presenting mEBAL, we also include preliminary experiments on: \textit{i)} eye blink detection using Convolutional Neural Networks (CNN) with the facial images, and \textit{ii)} attention level estimation of the students based on their eye blink frequency.  
\end{abstract}




\begin{teaserfigure}
  \centering
  \includegraphics[width=0.95\textwidth]{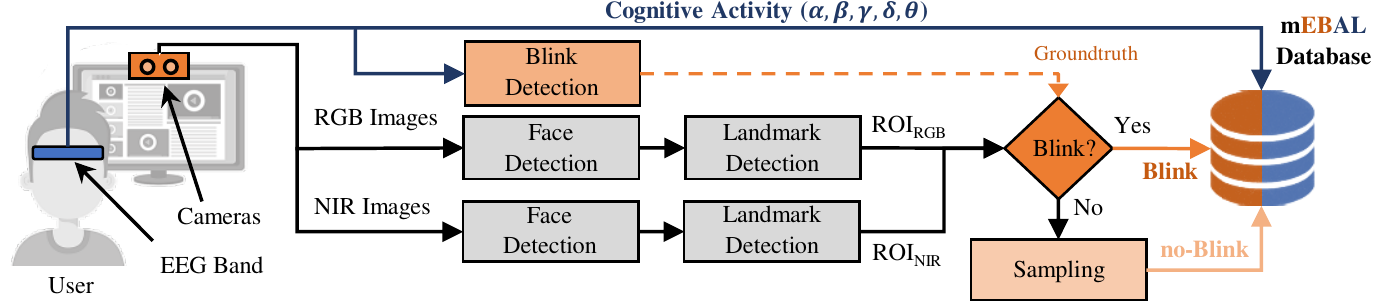}
  \caption{Diagram of the acquisition framework of mEBAL.}
  \Description{Block diagram of the acquisition framework of mEBAL.}
  \label{fig:teaser}
\end{teaserfigure}

\maketitle

\section{Introduction}

The importance of virtual education platforms has increased significantly in the last 10 years \cite{chen2018research} and the COVID-19 outbreak in 2020 has strengthen that importance. With a large percentage of the academic institutions around the world now in lockdown, virtual education has temporally replaced traditional education to a very large extent.

E-learning is not only useful in exceptional cases with an imposed social distancing, but also in many other scenarios in which traditional education is limited, as it can provide worldwide access, flexible schedule, and personalized learning strategies. However, e-learning also presents some challenges compared to the traditional face-to-face education, e.g.: the absence of a direct contact between teachers and students and the difficulties to certify the authorship during online evaluations. 

E-learning platforms allow to capture student information to create personalized environments whose contents and methodologies can be adapted dynamically to the different needs of each student. Information such as the performance on questions, the time necessary to perform the different tasks, the emotional state \cite{shen2009affective}, or the the heart rate \cite{javier2020comparative} can be used to understand the student behavior and conditions \cite{hern2020heart}. 

Undoubtedly, e-learning platforms will benefit significantly by exploiting the attention level of the student \cite{pengpredicting}. This could be used to: \textit{i)} adapt dynamically the environment and content \cite{nappi2018context,2018_INFFUS_MCSreview2_Fierrez} based on the attention level, and \textit{ii)} improve the educational materials and resources with a posterior analysis of the e-learning sessions (e.g. detecting the type of contents more appropriate for a specific student). 

Since the 70s there are studies relating the eye blink rate with cognitive activity like attention \cite{holland1972blinking,bagley1979effect}. The studies suggest that lower eye blink rates can be associated to high attention periods while higher eye blink rates are related to low attention levels. Therefore, in this context, automatic eye blink detection can be a tool for estimating the attention level of the students and improving e-learning platforms. 

\begin{table}[t]
  \caption{Existing databases for Eye Blink detection.}
  \label{tab:datasets}
  \renewcommand{\arraystretch}{1}
  \begin{tabular}{c c c c c c c c}
    \toprule
   \bf{Ref.} & \bf{Year} & \bf{Blinks}  & \bf{Users} & \bf{Resol.}  & \bf{Att.} & \bf{Sensors}\\
    \midrule
    \cite{Talkingface}& NA &  $61$ & $1$ & $720\times576$  & No & $1$
    \\
    \cite{pan2007eyeblink} & 2007 & $255$ & $20$ & $320\times240$ & No  & $1$
    \\
    \cite{drutarovsky2014eye} & 2014 & $353$ & $4$ & $640\times480$ & No  & $1$
    \\
    \cite{radlak2015silesian} & 2015 & $300$ & $5$ & $640\times480$  & No  & $1$
    \\
    \cite{hu2019towards} & 2019 & $381$ & $172$ & $1280\times720$  & No  & $1$\\ 
    \textbf{Ours} & \textbf{2020}& \bf{3000} & $\textbf{38}$ & \bf{1280 $\times$ 720} & $\textbf{Yes}$  & $\textbf{3}$\\
    \bottomrule
  \end{tabular}
\end{table}

 Similar rationale motivates also the usage of eye blink detection in many other problems where knowing the cognitive activity is useful, e.g.: driver fatigue detection \cite{bergasa2006real,hernandez2019quality}, lie detection \cite {mann2002suspects, leal2008blinking}, detection of mild cognitive impairment \cite{ladas2014eye}, face anti-spoofing \cite{pan2007eyeblink,2019_BookPAD2_IntroFacePAD_JHO,galbally14reviewAntispoofingFace}, dry eye syndrome recovery \cite{rosenfield2011computer}, human-computer interfaces \cite{acien2020smartphone} that ease communication for disabled people \cite{torricelli2009adaptive}, fake news and DeepFakes detection \cite{jung2020deepvision, DeepFakes_blinking2018,tolosana2020deepfakes}, and activity recognition \cite{ishimaru2014blink}. Even though our work has been developed with an e-learning application in mind, our research and resources (in particular mEBAL) can be very useful in all these related problems.

In the present work, two different approaches are used to detect eye blinks: \textit{i) EEG-based detection,} using electroencephalography signals acquired with a head band. This detection is used to create a candidate list of eye blinks that will be then manually validated to create a groundtruth. And \textit{ii) Image-based detection,} using the face images acquired using RGB and NIR cameras. This detection comprises face detection, landmark detection to locate the eye region, and eye blink detection. Fig. \ref{fig:teaser} shows the acquisition framework of mEBAL and the proposed eye blink detectors. The image-based detector could be used in applications where only a webcam is available. The main contributions of this study are:
\begin{itemize}
\item A new multimodal database for Eye Blink detection and Attention Level estimation: mEBAL\footnote{\url{https://github.com/BiDAlab/mEBAL}}. The database comprises data from 38 students and it includes: $3$,$000$ blink samples ($378$,$000$ frames in total), $3$,$000$ no-blink samples ($378$,$000$ frames in total), and the cognitive activity of the students. This database is eight times larger than existing eye blink databases and it is unique in its multimodal nature.
\item An image-based eye blink detector trained over the proposed mEBAL and evaluated over another public eye blink database \cite{hu2019towards}, considering in-the-wild scenarios. This detector consists of Convolutional Neural Networks (CNN).  
\item A preliminary experiment evaluating the eye blink detector as an estimator of the level of attention from RGB face images.  
\end{itemize}

The rest of the paper is organized as follows: Section \ref{related_works} summarizes works related to eye blink detection. Section \ref{MuBA_database} presents our database. Section \ref{algorithm} describes the baseline eye blink detector developed for our experiments. Section \ref{experiments_results} analyzes the results obtained on eye blink detection and attention level estimation. Finally, remarks and future work are drawn in Section \ref{conclusion}.

\begin{table}
 \renewcommand{\arraystretch}{1}
  \caption{Sensors included in the mEBAL framework.}
  \label{tab:sensors}
  \renewcommand{\arraystretch}{1}
  \begin{tabular}{c c l}
    \toprule
    \textbf{Sensors} & \textbf{Sampling Rate} & \textbf{Features}\\
    \midrule
       1 RGB Camera & 30 Hz&MP4  with codec H.264\\ 
       2 Infrared Cameras & 30 Hz&MP4  with codec H.264 \\
       %1 Depth camera& 30 Hz&MP4  with codec H264\\
       EEG Band & 1 Hz & \parbox{3cm}{$\alpha, \beta, \gamma, \delta, \theta$}\\
  \bottomrule
\end{tabular}
\end{table}

\section{Related Works}\label{related_works}

One of the challenges to train robust eye blink detectors is the absence of large-scale public databases. Table \ref{tab:datasets} summarizes the most popular eye blink detection databases in the literature. As can be seen, all available databases comprise only a few hundred samples. The main differences lie in the number of users and acquisition conditions. While databases such as \cite{drutarovsky2014eye,Talkingface,pan2007eyeblink}  present controlled environments, recent works explore the detection of eye blinks in unconstrained scenarios \cite{hu2019towards}. 

Regarding the detection methods, they can be divided into: \textit{i)} \textit{motion-based} \cite{morris2002blink, drutarovsky2014eye}, which exploit the dynamic information around eye blink events; \textit{ii)} \textit{appearance-based} \cite{dalal2005histograms, soukupova2016eye}, which process the images and extract features related to the texture and shape of the eyes; and \textit{iii)} \textit{mixed approaches} based on the combination of image and temporal information \cite{hu2019towards}.  

To the best of our knowledge, our contributed mEBAL database is the first one that provides both eye blink images and cognitive activity information. 

%We consider a controlled acquisition scenario, similar to e-learning scenarios. Fig. \ref{fig:Examples_eyes} shows example images included in mEBAL. Images were acquired using both NIR and RGB cameras. In total, 6,000 samples are included (3,000 blink and no-blink samples each), being the largest database to date (eight times larger than previous ones). The database include the cognitive activity of all 38 students during the whole acquisition period (varying from 15 to 30 minutes). It is important to highlight the challenging of the task as in some non-eye blink examples the eyes can seem to be closed due to the gaze orientation, e.g., when looking at the keyboard instead of the cameras. Also, other aspects of variability such as the user position, presence of glasses, or changes in the illumination are considered in our database. 

\section{mEBAL: Multimodal Framework for Eye Blink and Attention Level}\label{MuBA_database}
\subsection{Acquisition Sensors}

We designed a multimodal acquisition framework to monitor cognitive and eye blink activity during the execution of online tasks (see Fig. \ref{fig:teaser}) based on the edBB platform for remote education assessment \cite{hernandez2019edbb}. The following sensors are considered in the acquisition setup (see Table \ref{tab:sensors}):

\begin{itemize}
\item An \textbf {EEG headset} by NeuroSky\footnote{\url{https://store.neurosky.com/pages/mindwave}} that captures 5 channels of electroencephalographic information ($\alpha, \beta, \gamma, \delta, \theta$). These signals provide temporal information related to the cognitive activity of the student. The sensor also provides a temporal sequence with the eye blink strength. The sampling rate of the band is $1$ Hz. The EEG band is used to capture the cognitive activity of the student and the eye blink candidates that will be used as groundtruth (as we will see later, some eye blinks must be discarded as false positives).
\item An \textbf {Intel RealSense} (model D435i), which comprises $1$ RGB and $2$ NIR cameras. An average blink takes $100$ms-$400$ms according to the Harvard Database of Useful Biological Numbers \cite{schiffman1990sensation}. The  Intel RealSense sensor is configured to $30$ Hz (one frame every $33$ms) and $1280\times720$ resolution. Therefore, an eye blink can take between $3$ to $13$ frames.  

\end{itemize}

\begin{figure}[t]
  \centering
  \includegraphics[width=\linewidth]{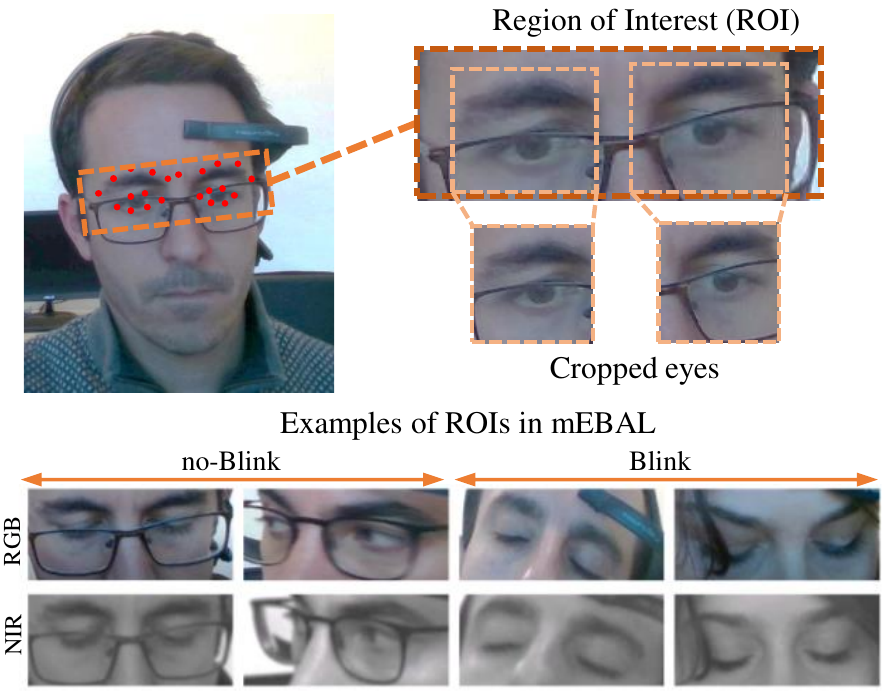} %./Figures/Figure_example_opened_closed_eyes3.jpg
  \caption{Landmark detection and ROI extraction. Example images included in mEBAL. Note that in some no-blink cases the eyes seem to be closed due to the gaze orientation.}
  \Description{Examples of opened/closed eyes on mEBAL}
   \label{fig:Examples_eyes}
\end{figure}

\subsection{Description of the mEBAL Database} \label{database}

Students performed 8 different tasks categorized in the following three groups: \textit{i) enrollment form:} name and surname, ID number, nationality, e-mail address, etc. (low level of attention is expected); \textit{ii) writing questions:} these questions are oriented to measure the students' cognitive abilities under different situations such as solving logical problems, describing images, crosswords, finding differences, etc. (increasing level of attention is expected); and \textit{iii) multiple choice questions:} aimed to detect the students' attention and focus levels (high level of attention is expected).

mEBAL comprises a total of $6$,$000$ samples divided in two halves (blink and no-blink) from both eyes acquired with $1$ RGB and $2$ NIR cameras. Each sample comprises $21$ frames (around $600$~ms.) for a total number of images of $756$,$000$ ($6$,$000\times21\times2\times3$). Aspects such as the user position and changes in the illumination  were considered during the acquisition in order to simulate realistic e-learning scenarios. 11 out of the 38 students used glasses.

The mEBAL dataset was obtained from the raw data provided in the edBBdb \cite{hernandez2019edbb}. The eye blink and attention level information was labelled following a semi-supervised method. First, eye blink candidates were selected using the EEG band signals (eye blink strength is an attribute provided by the EEG band SDK). Second, we made a manual refinement of the eye blink samples detected by the band to eliminate false positives. Once the eye blink samples were validated, we stored the 10 frames previous and posterior to the eye blink event ($21$ frames in total for each eye blink). These frames can be used to exploit the temporal information proposed in some approaches of the literature \cite{soukupova2016eye}. Finally, we  used facial landmark detection \cite{dlib} to track the eye position (see Fig. \ref{fig:Examples_eyes}). Entire face images, eye bounding boxes, and the cropped eyes are provided in our contributed mEBAL database. Additionally, we include the cognitive temporal signals $\alpha, \beta, \gamma, \delta, \theta$ provided by the EEG band. As the number of no-blink images is much larger than the blink images, we subsampled the no-blink images to obtain the same number of samples per class: Blink and no-Blink (see Fig.~\ref{fig:teaser}). 

\section{Eye Blink Detection Algorithm} \label{algorithm}

Inspired in the popular VGG16 architecture \cite{simonyan2014very}, we propose an eye blink detector based on a CNN trained from scratch. The proposed network comprises an input layer of $50 \times 50$ size, followed by $3$ convolutional layers with ReLU activation ($32/32/64$ filters of size $3\times3$), with $3$ max pooling layers between them, a dense layer of $64$ units with ReLU activations, and a final output layer with one unit (sigmoid activation). Also, we use dropout ($0.5$) to reduce overfitting. The batch size is set up to $50$. Adam optimizer is considered with default parameters ($0.001$ learning rate). The network is trained as a binary classifier (eyes open or closed), using the mEBAL subset of RGB cropped eyes (see Fig.~\ref{fig:Examples_eyes}).      

\begin{table}[t]
  \caption{Eye Blink detection results on HUST-LEBW dataset.}
  \label{tab:Comparison}
 
  \begin{tabular}{c c c c c}
  %\setlength\extrarowheight{3pt}
    %\toprule
    \hline
     \multirow{1}{1.5cm}{ \centering \bf{Method}}&  \multirow{1}{1.5cm}{\centering \bf{Eye}} &  \multirow{1}{1cm}{\centering \bf{Recall}} & \multirow{1}{1cm}{\centering \bf{Precision}} & \multirow{1}{1cm}{\centering \bf{F1 }} \\
    %\midrule
    \hline \hline
     \multirow{2}{1.3cm}{\centering \cite{morris2002blink}} & \multirow{2}{1cm}{ \centering Left\\Right}
     & \multirow{2}{1cm}{\centering 0.0164 \\  0.0159}
     & \multirow{2}{1cm}{\centering 0.6667 \\ \textbf{1.0000}}
     & \multirow{2}{1cm}{\centering 0.0320 \\ 0.0313} 
 
 \\ \\ \hline
 \multirow{2}{1cm}{\centering \cite{chau2005real}} & \multirow{2}{1cm}{ \centering Left\\Right}
     & \multirow{2}{1cm}{\centering  0.0164 \\ 0.0000}
     & \multirow{2}{1cm}{\centering \textbf{1.0000} \\   0.0000}
     & \multirow{2}{1cm}{\centering  0.0323 \\  0.0000} 
 \\ \\ \hline
    \multirow{1}{1cm}{\centering \cite{tabrizi2008open}} & \multirow{1}{1cm}{ \centering 2 eyes}
     & \multirow{1}{1cm}{\centering 0.0714 }
     & \multirow{1}{1cm}{\centering 0.4500 }
     & \multirow{1}{1cm}{\centering   0.1233 } 
 \\   \hline
 \multirow{2}{1cm}{\centering \cite{drutarovsky2014eye}} & \multirow{2}{1cm}{ \centering Left\\Right}
     & \multirow{2}{1cm}{\centering  0.0574 \\ 0.0317}
     & \multirow{2}{1cm}{\centering 0.4118 \\    0.3077}
     & \multirow{2}{1cm}{\centering  0.1007 \\  0.0576} 
\\ \\ \hline
\multirow{2}{1cm}{\centering \cite{soukupova2016eye}} & \multirow{2}{1cm}{ \centering Left\\Right}
     & \multirow{2}{1cm}{\centering  0.3607 \\ 0.3016}
     & \multirow{2}{1cm}{\centering 0.6471 \\    0.5758}
     & \multirow{2}{1cm}{\centering  0.4632 \\  0.3958} 
\\ \\ \hline
\multirow{2}{1cm}{\centering \cite{hu2019towards}} & \multirow{2}{1cm}{ \centering Left\\Right}
     & \multirow{2}{1cm}{\centering  0.5410 \\ 0.4444}
     & \multirow{2}{1cm}{\centering 0.8919 \\   0.7671}
     & \multirow{2}{1cm}{\centering  0.6735 \\  0.5628} 
\\ \\ \hline  
\multirow{2}{1cm}{\centering \textbf{Ours}} & \multirow{2}{1cm}{ \centering Left\\Right}
     & \multirow{2}{1cm}{\centering  \textbf {0.9603} \\ \textbf {0.7950}}
     & \multirow{2}{1cm}{\centering 0.6080 \\   0.7348}
     & \multirow{2}{1cm}{\centering \textbf {0.7446} \\  \textbf {0.7637}} 
    %\bottomrule
    \\ \\ \hline
  \end{tabular}
\end{table}

\section{Experiments and Results}\label{experiments_results}

\begin{figure*}[ht]
  \centering
  \includegraphics[width=\linewidth]{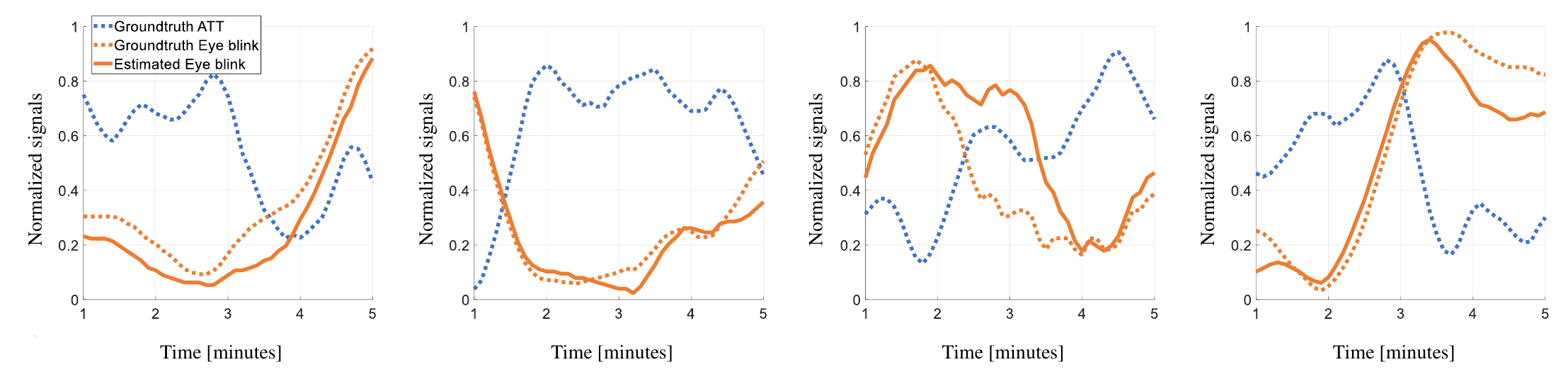}
  \caption{Normalized Attention Level, estimated and groundtruth Eye Blink rates of 4 different students.}
  \Description{Temporal evolution of the attention, estimated eye blink and groundtruth eye blink.}
   \label{fig:attention}
\end{figure*}

\subsection{Experimental Protocol}

The evaluation is performed over the public HUST-LEBW benchmark for eye blink detection presented in \cite{hu2019towards}. Note that the mEBAL database (used for training our blink detector) was obtained in a controlled environment, while the HUST-LEBW dataset (used for testing our blink detector) was obtained in the wild. This will allow to measure the generalization ability of the proposed eye blink detector to unseen scenarios \cite{2018_TIFS_SoftWildAnno_Sosa,2018_IntelSys_Trends_Proenca}.

The HUST-LEBW dataset includes $381$ eye blink and $292$ no-blink samples. Each sample comprises $13$ frames. All $13$ frames are processed with the CNN proposed in Section~\ref{algorithm}, which generates for each input image an eye blink strength score. Among the $13$ scores obtained for a sample (one per frame), the maximum is selected to represent the sample score. The decision threshold is fixed to the point in which the False Positive and False Negative rates in blink detection are equal.

\subsection{Eye Blink Detection Results}

Table \ref{tab:Comparison} presents the results and comparison with previous approaches evaluated over the same HUST-LEBW Dataset \cite {hu2019towards,soukupova2016eye,tabrizi2008open,chau2005real,morris2002blink, drutarovsky2014eye}. The results show how our method outperforms state-of-the-art eye blink detection algorithms for Recall and F1 metrics. There is an important difference between left and right eyes in terms of performance. This difference is caused by the characteristics of the HUST-LEBW dataset (e.g., head orientation in the samples).

Note that the approach proposed in \cite {hu2019towards} was developed using the training set provided with HUST-LEBW. This training set comprises images with similar characteristics as those used in the evaluation set. Our results were obtained training with mEBAL, acquired under controlled conditions. Thus, the high performance obtained with our method demonstrates the good generalization capacity of our approach to unseen scenarios. These results demonstrate the potential of mEBAL to train a new generation of eye blink detectors.

\subsection{Attention Level Estimation Results}

In this section, we analyze the relationship between  the eye blink rate and the attention level estimated by the algorithm provided with the EEG band. We present a preliminary experiment focused on sudden changes in the level of attention of students. According to the literature \cite{holland1972blinking,bagley1979effect}, we should observe  an inverse relationship between eye blink rate and level of attention. Fig. \ref{fig:attention} shows the attention level and eye blink estimation of $4$ different students during a period of $4$ minutes. We have averaged the level of attention for each $20$ seconds using an sliding window of $5$ seconds. The estimated and groundtruth eye blinks are calculated as bpm (blinks per minutes) using a sliding window of $5$ seconds again. All three signals were normalized using the \textit{min-max} technique \cite{jain2005score,2018_INFFUS_MCSreview1_Fierrez}. 

The results show a small difference between the eye blink detector and the groundtruth provided by the EEG band. These results demonstrate again the high accuracy of the trained image-based blink detector. Regarding the level of attention and the eye blink rate, in most instances, the level of attention and eye blink signals have a negative correlation, which is coherent with the literature. High peaks in the attention level are usually correlated with low eye blink rates and viceversa. However, other factors can affect the eye blink rates and more context is necessary to improve the estimation of the level of attention based on the eye blink rate.

\section{Conclusion}\label{conclusion}
This work has presented mEBAL, a new multimodal database for eye blink detection and attention level estimation, the largest one in the literature for research in these problems. This database improves previous databases in sensors (EEG band, NIR and RGB cameras) and samples: 6,000 samples in two halves (blink and no-blink) of both eyes for a total of 756,000 images.

Also, we have performed experiments to: \textit{i)} detect the eye blinks with Convolutional Neural Networks trained from scratch using RGB images, and \textit{ii)} predict the level of attention of students conducting various e-learning tasks based on their eye blink frequency.

The results achieved have proved that mEBAL can be used to train accurate eye blink detectors under realistic acquisition conditions. In fact, our results have outperformed the state of the art with a simple yet powerful CNN learning architecture, thanks mainly to the utility of our contributed database, which is well suited for data-driven eye blink detection approaches.

Future work should consider other recognition architectures better adapted to the eye blink detection problem (e.g., combining CNNs and Recurrent Neural Networks \cite{2018_INFFUS_MCSreview1_Fierrez,baltruvsaitis2018multimodal} to incorporate time information \cite{tolosana2020biotouchpass2}). The preliminary experiment carried out to measure the correlation between the attention level and the eye blink frequency has shown encouraging results and should be also analyzed in more depth. 

Even though our work has been developed with e-learning in mind \cite{hernandez2019edbb,hern2020heart}, the contributed resources and methods for eye blink detection can be very useful for other problems as well, e.g.: driver fatigue detection \cite{hernandez2019quality}, lie detection, DeepFakes detection \cite{jung2020deepvision}, face anti-spoofing \cite{2019_BookPAD2_IntroFacePAD_JHO}, human-computer interfaces \cite{acien2020smartphone}, and others.

\begin{acks}
This work has been supported by projects: PRIMA (ITN-2019-
860315), TRESPASS-ETN (ITN-2019-860813), IDEA-FAST (IMI2-2018-15-two-stage-853981), BIBECA
(RTI2018-101248-B-I00 MINECO-FEDER), and edBB (Universidad Autonoma de Madrid). Ruben Tolosana and postdoc support from CAM/FEDER. Roberto Daza is supported by a PhD FPI fellowship from MINECO-FEDER.
\end{acks}

\clearpage

\bibliographystyle{ACM-Reference-Format}
\balance
\bibliography{sample-base}

\end{document}